\documentclass[11pt]{scrartcl}
\usepackage[margin=1.3in]{geometry}
\usepackage{graphicx}

\usepackage[T1]{fontenc}
\usepackage[utf8]{inputenc}

\usepackage{xcolor}
\definecolor{gray}{gray}{0.4}

\usepackage[round,authoryear]{natbib}
\usepackage{expex}
\lingset{everygla=\it, aboveexskip=1em, belowexskip=1em}

\usepackage{hyperref}
\hypersetup{
  colorlinks=true,
  linkcolor=black,
  citecolor=black,
  urlcolor=black
}

\title{Grammaticality Judgments in Humans and Language Models:\\
Revisiting Generative Grammar with LLMs}

\author{
  Lars G.~Bagøien Johnsen\\[0.3em]
  \small National Library of Norway\\
  \small \texttt{lars.johnsen@nb.no}
}
\date{2025}

\begin{document}

\maketitle


\begin{abstract}
What counts as evidence for syntactic structure? In traditional generative grammar, systematic contrasts in grammaticality such as subject–auxiliary inversion and the licensing of parasitic gaps, are taken as evidence for an internal, hierarchical grammar. In this paper, we test whether large language models (LLMs), trained only on surface forms, reproduce these contrasts in ways that imply an underlying structural representation.

We focus on two classic constructions: subject–auxiliary inversion (testing recognition of the subject boundary) and parasitic gap licensing (testing abstract dependency structure). We evaluate models including GPT-4 and LLaMA-3 using prompts eliciting acceptability ratings. Results show that LLMs reliably distinguish between grammatical and ungrammatical variants in both constructions, and as such supports that they are sensitive to structure and not just linear order. Structural generalizations, distinct from cognitive knowledge, emerge from predictive training on surface forms, suggesting functional sensitivity to syntax without explicit encoding.
\end{abstract}

\section{Introduction}

LLMs are often described as next-word predictors trained on token sequences, in contrast to human language processing, which appears to rely on internalized syntactic structure. A core piece of evidence for this comes from subject–auxiliary inversion in yes/no questions: humans move the matrix auxiliary, not the linearly first one. Children, tellingly, never produce errors like *\emph{Had the boy who eaten the apples will go home?}, suggesting an implicit notion of subjecthood and constituent structure \cite{crain1987structure, ziv2025largelanguagemodelsproxies}.

A common misunderstanding is to treat next-token prediction as a description of an LLM’s internal representations. Yet training objectives dictate only how a model learns, not what kinds of structures it can deploy at inference time. BERT \footnote{While masked-LM objectives (e.g.\ BERT) have been central in NLP, several studies now suggest that such objectives may yield less stable global representations than autoregressive training \citep{}{TenneyEtAl2019, Manning2023LLM}. This may explain why structural generalizations are more consistently expressed in models trained with next-token prediction.}
 and GPT differ sharply in objective (masked-token prediction vs.\ autoregressive next-token prediction), but once an input sequence is given, both operate over high-dimensional states that support global, structure-sensitive computations. In this sense, next-token prediction is a \textit{surface projection} of a much richer representational geometry. The hierarchical behaviour observed in our experiments is therefore better understood as emerging from this geometry, rather than as a consequence of the superficial training objective alone.

We hypothesize that LLMs’ grammaticality judgments reflect structural generalizations derived from training data without presupposing cognitive knowledge \cite{chomsky1965aspects}. Linguistic theory infers constituent structures (e.g., noun phrases with heads and adjuncts like relative clauses) from human judgments to explain systematic contrasts, such as yes/no question formation. By positing operations like verb movement after the initial noun phrase, these structures are developed into various grammatical concepts like for example the syntactic functions of sentence subjects. Phrase structure trees, alongside dependency structures, have been central to generative grammar for over 50 years.  

The use of LLMs to probe syntactic structure recalls these earlier (and ongoing) practices in generative grammar, where introspective judgments were the primary data for linguistic theory. Note that introspection do not mean that structures as such are observed. While modern linguistics increasingly draws on sociolinguistic variation and corpus methods, the core argument for phrase structure remains rooted in explanatory adequacy: subject–auxiliary inversion or parasitic gap licensing are best accounted for by assuming constituent boundaries. As Jackendoff \cite{jackendoff2021meaning} has argued, even alternatives to hierarchical syntax ultimately rely on structured representations to explain linguistic behavior.

Given the role of grammaticality judgments, we want to put LLMs in the role of humans, and see if they exhibit the same behaviour with respect to key constructions, i.e. whether LLMs demonstrate behavior consistent with human-like structural sensitivity. If they reliably rate on a par with humans, we may infer that their internal mechanisms support some form of syntactic representation, and that it may be described in the same terms as we do for humans. Now, even though we do have more access to the the inner workings of LLMs, we will confine ourselves to talk about sentences, judgments and theoretical constructs built out of sentences, like grammars. Our study partially complements recent graded judgment studies \cite{qiu2024evaluating, studenikina2025gradual} showing LLMs' alignment with human intuitions.

The question of whether syntactic knowledge is innate or acquired through interaction with linguistic input has long been debated, notably in the exchange between Chomsky and Piaget \cite{piattelli1980language}. While Chomsky argued for an innate Universal Grammar to account for human linguistic competence, Piaget maintained that linguistic knowledge emerges through general cognitive development and interaction with the environment. Our findings contribute to this debate from a different angle: if large language models, trained solely on surface forms, exhibit a behavior similar to human-like grammatical knowledge, then such competence may emerge from statistical learning alone, without presupposing a biologically specified grammar.

In this light, LLMs serve as empirical tools for probing grammatical patterns, complementing human judgments, extending the domain of grammatical evidence by offering parallel patterns of sensitivity to structure. Our test items centered on gaps, licensing and movement, can be treated as epistemic objects in Rheinberger’s sense \cite{rheinberger1997epistemic}, carrying theoretical significance while remaining stable across frameworks. Even if the terminology—“parasitic gap,” “that-trace effect”—is theory-laden, the acceptability contrasts they describe provide a shared descriptive ground.

\section{LLMs as Theoretical Proxies}

In \cite{ziv2025largelanguagemodelsproxies}, the authors argue that large language models should not be viewed as full-fledged theories of human linguistic cognition, but rather as \textit{proxies}: tools that allow researchers to test which linguistic generalizations can be acquired from input alone. They distinguish between:

\begin{itemize}
  \item \textbf{The LLM Theory}: the idea that LLMs represent complete models of human linguistic competence;
  \item \textbf{The Proxy View}: a more cautious stance, using LLMs to evaluate the plausibility of different hypotheses about what is learnable from linguistic experience.
\end{itemize}

We adopt and extend the proxy view. Our experiments show that LLMs like GPT-4 and LLaMA-3 can successfully distinguish grammatical from ungrammatical forms in constructions that have played central roles in positing linguistic structures. These findings suggest that such structural generalizations may be learnable from form-based input alone, without explicit instruction or innate bias. In this sense, LLMs serve not just as tools for language generation, but as \textit{empirical instruments for testing syntactic hypotheses}. Their graded performance across constructions also helps us identify which phenomena are more easily learned from distributional data (e.g., inversion, parasitic gaps), and which remain challenging (e.g., across-the-board extraction or definiteness effects). This supports a view of syntax as an emergent property that is statistically induced rather than hard-coded.

\paragraph{On the nature of structure.}
In generative grammar, structure is a theoretical construct used to explain and formalize the linguistic competence or knowledge attributed to speakers. The structures posited (e.g., phrase markers, movement chains) are not themselves observed, but inferred from patterns in acceptability judgments. This distinction becomes crucial when evaluating LLMs. While LLMs may exhibit behaviors that align with structured representations, we should be careful not to reify structure as something internal to the model. Instead, their behavior offers evidence for structured output or structure-sensitive performance, which may or may not reflect structured knowledge in the human sense. Our position aligns with the proxy view: we use LLMs to test which structural generalizations can be induced from form alone—not to claim that LLMs possess syntactic competence in the same cognitive sense as humans.
Importantly, it is not the LLMs themselves, as instances of artificial intelligence, that explain these effects, but the structure latent in the training data. The models reflect, rather than invent, the distributional patterns present in language use. Their ability to act as theoretical proxies depends on the richness and regularities of the data they are trained on.

\paragraph{Relation to earlier work.} Earlier studies have examined the performance of LLMs on long-distance dependencies and extraction phenomena. In \cite{chowdhury-zamparelli-2018-rnn} it was demonstrated that RNN-based models could capture some ungrammaticality effects due to island violations, but lacked robust generalization, while the study \cite{warstadt-etal-2020-blimp} constructed and used a large dataset (BLiMP) to assess linguistic knowledge in LLMs and found partial alignment with syntactic constraints. Our work continues this line but focuses on constructions that are specifically diagnostic of structure (e.g., filler-gap dependencies, auxiliary movement) and evaluates whether current models capture these phenomena.

\paragraph{Parasitic gaps and ATB.} Some analyses have noted structural parallels between parasitic gaps and across-the-board (ATB) extraction. In particular, when the parasitic gap appears in an adjunct clause that is semantically or structurally parallel to the matrix clause, the dependency may resemble coordination. This has led to proposals, such as \cite{postal1994parasitic} and \cite{nunes2001linearization}, that treat certain parasitic gaps as derivationally similar to ATB, relying on mechanisms like multidominance or shared movement. While our results treat these constructions separately, the theoretical connection may explain why both phenomena challenge LLMs differently: parasitic gaps are handled robustly, while ATB extractions yield more variable judgments.

In this study, we use LLMs as instruments for testing which syntactic distinctions can be learned from data alone. By comparing model ratings to human acceptability judgments, we assess whether constructions that motivated generative structure, such as parasitic gaps and subject-auxiliary inversion, are recoverable from input distributions. If so, this supports a view of grammar as an emergent property of linguistic exposure rather than an innate module. The following section details our experimental setup.

\section{Method}

To evaluate structural sensitivity in LLMs, we designed a test suite of 80 sentence sets in English and Norwegian, targeting three syntactic constructions central to generative grammar:

\begin{itemize}
\item \textbf{Parasitic gaps}, including diagnostic “plugging” with pronouns
\item \textbf{Across-the-board (ATB) extraction} from coordinated clauses
\item \textbf{Subject–auxiliary inversion} in embedded yes/no questions
\end{itemize}

Each item consisted of minimal pairs differing only in grammaticality. We presented these to instruction-tuned LLMs (GPT-4 and LLaMA-3) using prompts that elicited acceptability ratings on a 1–5 scale. This setup allowed us to measure alignment with human judgments and structural preferences across languages.

Each construction was presented as minimal pairs of grammatical and ungrammatical sentences. Instruction-tuned models (GPT-4 via OpenAI API, and Meta's \texttt{meta/llama3-405b-instruct-maas} via Hugging Face) were asked to rate sentence acceptability on a 1--5 scale. This allowed us to measure how closely model preferences align with native speaker intuitions and whether language-specific variation is preserved.

These constructions were selected because they are classically used to argue for deep syntactic representations in human language. Parasitic gap items were split into two types based on gap order: \textbf{LP} (licensing gap before parasitic) and \textbf{PL} (parasitic before licensing). Each type included four variants:

\begin{itemize}
\item \textbf{0}: both gaps unfilled (grammatical)
\item \textbf{1}: first gap filled (ungrammatical)
\item \textbf{2}: second gap filled (ungrammatical)
\item \textbf{3}: both gaps filled (ungrammatical)
\end{itemize}

We avoided textbook phrasing by varying verbs and lexical content while preserving syntactic structure. For example:

\pex
\a LP-0 Which organization did you donate money to because you believed in?
\a LP-1 Which organization did you donate money to it because you believed in?
\a LP-2 Which organization did you donate money to because you believed in it?
\a LP-3 Which organization did you donate money to it because you believed in it?
\xe

and:

\pex
\a PL-0 The German Shepherd is the kind of dog which everyone who owns often takes to the vet.
\a PL-1 The German Shepherd is the kind of dog which everyone who owns it often takes to the vet.
\a PL-2 The German Shepherd is the kind of dog which everyone who owns often takes it to the vet.
\a PL-3 The German Shepherd is the kind of dog which everyone who owns it often takes it to the vet.
\xe

\paragraph{Across-the-board extraction.}
ATB items tested whether the models respected the parallel movement between coordinated clauses. Each item included a grammatical baseline and an ungrammatical variant with asymmetrical extraction or plugging. For example:

\pex
\a Which book did John read \_ and Mary review \_?  (grammatical)
\a *Which book did John read it and Mary review \_? (ungrammatical)
\xe

These constructions are sensitive to structural symmetry and coordination constraints, making them useful diagnostics for constituent structure.

\paragraph{Subject–auxiliary inversion.}
To probe hierarchical structure, we included embedded yes/no questions where only the matrix auxiliary may undergo inversion. Ungrammatical examples fronted embedded auxiliaries, violating constituent boundaries. For example:

\pex
\a Will the boy who is crying \_ go outside? (grammatical)
\a *Had the boy who is crying will go outside?  (ungrammatical)
\xe

Correct responses depend on recognizing the boundary between matrix and embedded clauses, a key test of structural awareness.

\section{Results}

Instruction-tuned LLMs, including GPT-4 and \texttt{meta/llama3-405b-instruct-maas}, show high sensitivity to syntactic contrasts, with grammatical sentences typically rated 4–5 and ungrammatical ones 1–2. In our current dataset of 80 sentence sets (English and Norwegian), GPT-4 achieves near-perfect accuracy on English parasitic gaps and inversion (100\%), with slightly lower performance on ATB (83\%). For Norwegian, accuracy remains high on inversion (78\% perfect match) but drops for ATB (29\%). LLaMA follows a similar pattern with somewhat greater variability, especially in Norwegian.

These results indicate strong structural sensitivity in constructions involving hierarchy and licensing, but reduced robustness when symmetry or coordination is required. The following sections break down performance by construction type.

Overall, results indicate that LLMs—particularly instruction-tuned ones—capture core syntactic generalizations and mirror human-like patterns of grammatical reasoning.

\subsection{Parasitic Gaps}

Parasitic gaps have been central in debates about syntactic structure and innateness \cite{engdahl1983parasitic, taraldsen1981extraction, chomsky1982some}. They require the presence of a licensing gap—typically from wh-movement—before.

\paragraph{Robust alignment on licensing violations.}
To test structural sensitivity, we examined sentences where the licensing gap was filled (e.g., \textit{*Which organization did you donate money to it because you believed in?}), which are ungrammatical under standard analyses.

\begin{figure}[h]
  \centering
  \includegraphics[width=0.8\linewidth]{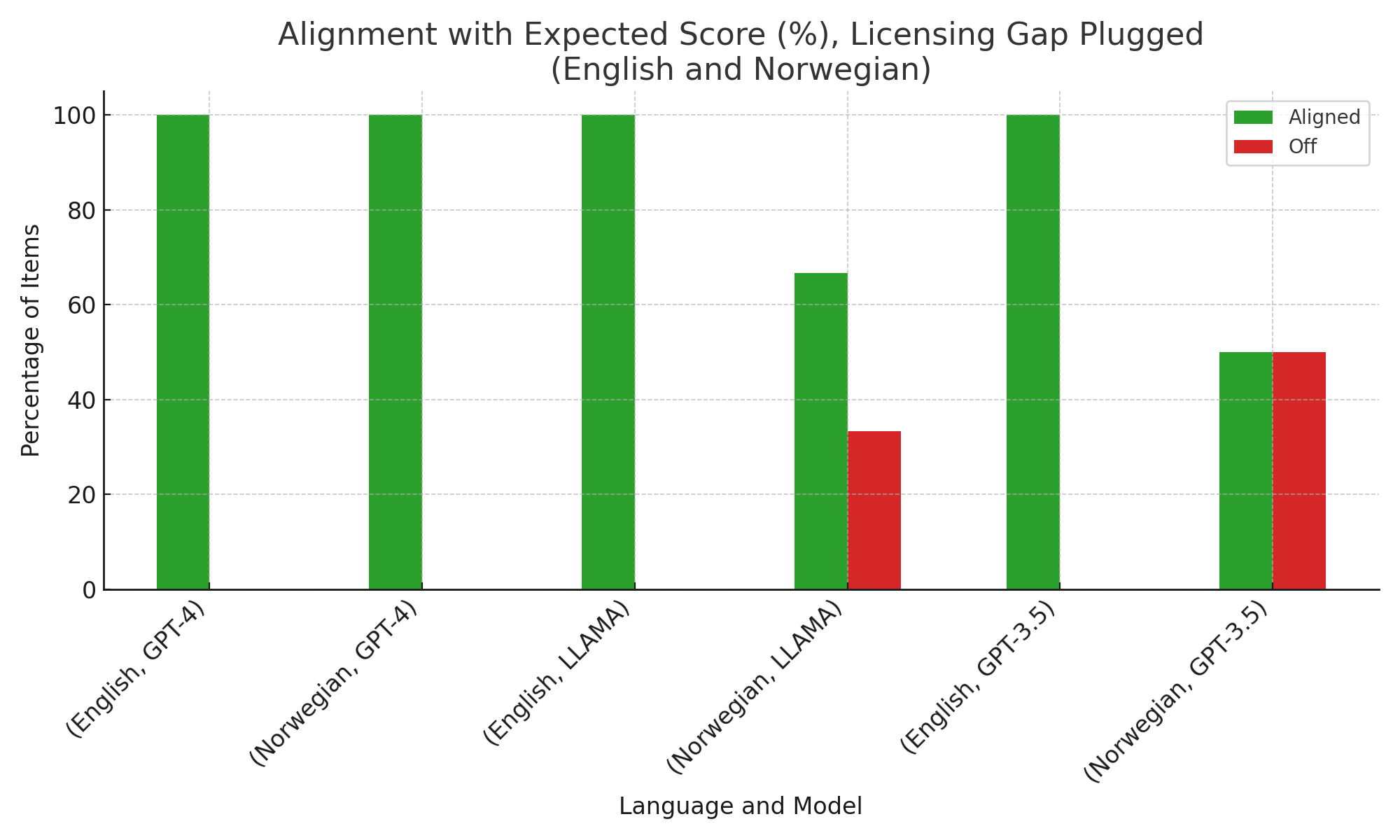}
  \caption{Accuracy of LLM judgments on parasitic gap constructions, by language and model.}
  \label{fig:licensing_gap_alignment_by_lang_model}
\end{figure}

Figure~\ref{fig:licensing_gap_alignment_by_lang_model} shows that all models rate these constructions with high alignment to expected judgments—especially GPT-4, which reaches near-perfect performance in English and strong alignment in Norwegian. The majority of model responses fall within one point of the expected value, demonstrating fine-grained grammatical awareness rather than binary pattern matching. This suggests that LLMs treat such constructions as structurally ill-formed, in line with hierarchical licensing requirements.
Most responses deviated by no more than one point from the expected judgments, which was always either 5 (fully grammatical) or 1 (strongly ungrammatical). This pattern suggests that the models assign graded well-formedness even to rare constructions, and that their structural sensitivity is not tied to linear order. Rather, they appear to encode licensing conditions and dependency constraints in a way that reflects island effects and constituent structure \cite{ross1967constraints, chomsky1982some}.

\begin{figure}[h]
  \centering
  \includegraphics[width=0.8\linewidth]{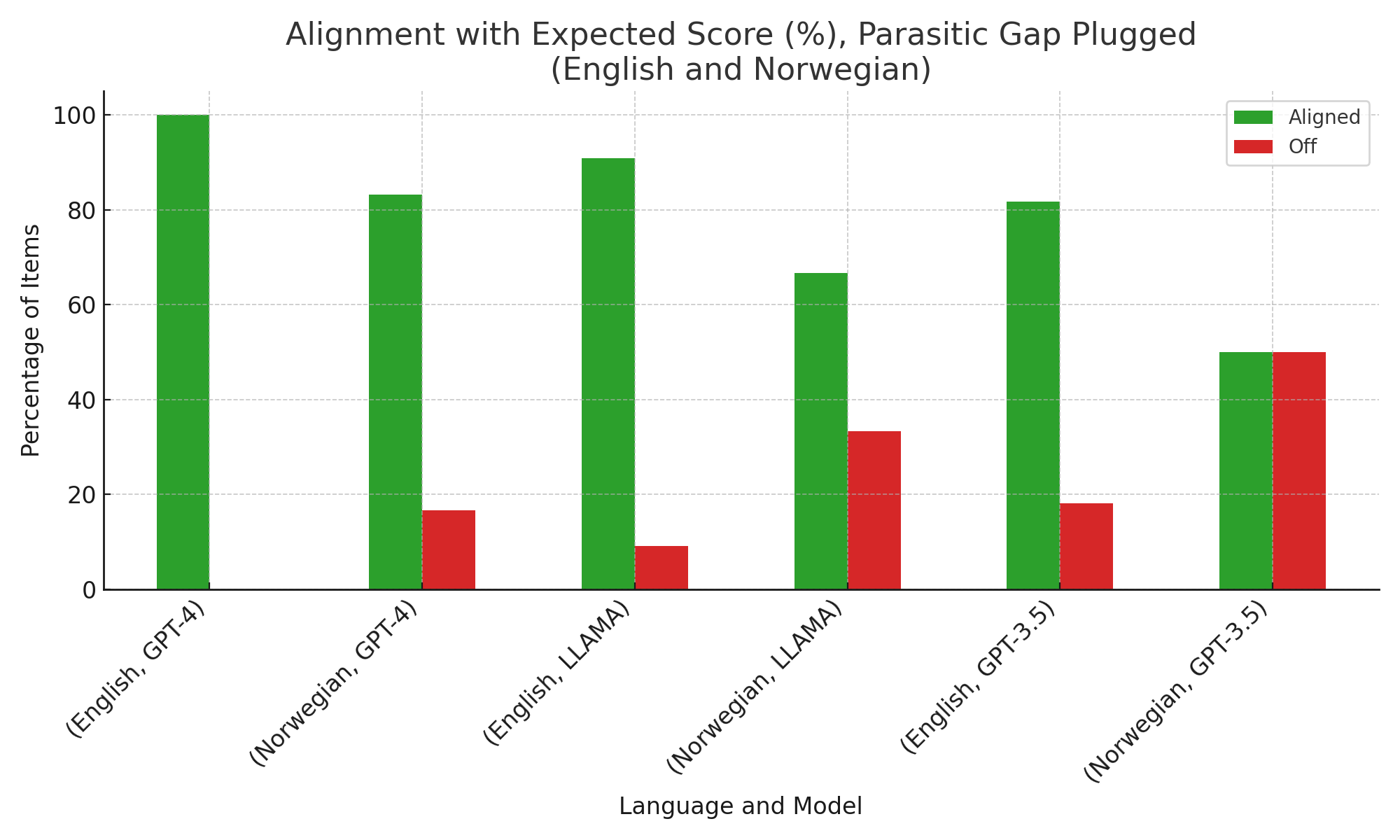}
  \caption{Accuracy of LLM judgments on parasitic gap constructions, by language and model.}
  \label{fig:parasitic_gap_alignment_by_lang_model}
\end{figure}

\paragraph{Parasitic gap sensitivity.}
We also examined sentences where the parasitic gap was “plugged” by a pronoun (e.g., \textit{*Which organization did you donate money to because you believed in it?}). These should in general be judged to be well formed sentences. Figure~\ref{fig:parasitic_gap_alignment_by_lang_model} shows fairly strong alignment between model judgments and expected scores, particularly in English. GPT-4 again shows the highest consistency, while LLaMA and GPT-3.5 display somewhat reduced performance in Norwegian. This may be due to the thematic complexity of some test items where several Norwegian examples involve semantically odd fillers (e.g., \textit{Her er himmelen mannen som ville klippe den opp prøver å knuse i tusen biter}), reminiscent of “colorless green ideas” constructions. This may point to the way LLMs build structure based on attention weights, but we leave this as an open question.

Despite this, all models penalize the plugged licensing gap variants appropriately in most cases, confirming that their representations go beyond surface form and reflect licensing relationships between gaps. The behaviour of LLMs on this construction is not easily reduced to locality of the verbs and the sequencing of gaps, in one the \textbf{LP} type the licensing gap comes before the parasitic gap, while in \textbf{PL} type it is the other way around. This further supports the claim that LLMs demonstrate behavior consistent with structural constraints, not readily reducible to linear order. Although exact matches are not always achieved, most ratings deviate by at most 1 point from the expected value, see again (Figure~\ref{fig:licensing_gap_alignment_by_lang_model}). 

\subsection{Across-the-board (ATB) Extraction}

ATB extraction involves coordinated structures where a single wh-element is interpreted as a gap in both conjuncts.

\pex
\a Which book did John read \_ and Mary review \_? \hfill (grammatical)
\xe

Acceptability depends on symmetrical extraction from parallel positions. GPT-4 matched expected judgments on 83\% of English ATB items and 29\% of Norwegian ones, with an average deviation of 0.8 and 2.0 points respectively. LLaMA followed a similar pattern, deviating on average by 0.5 points in English and 1.86 in Norwegian. These results contrast sharply with the near-perfect alignment observed for parasitic gaps, suggesting that ATB structures pose a greater challenge.

This discrepancy may reflect the complexity of coordinating symmetrical movement across clauses—a property not easily reducible to surface order or local dependencies. It also invites speculation on how structure is represented in LLMs: if syntactic behavior is attention-guided, then structures requiring structural balance rather than dependency chains may be harder to capture. Whereas parasitic gaps involve licensing relationships, ATB extraction requires a kind of abstract mirroring—possibly less salient in the input distribution. This may hint at an attention-based bias toward contentful elements over structural parallelism, though further probing is needed.

\subsection{Embedded Yes/No Questions and Inversion}

Subject–auxiliary inversion has long been used to motivate structural hierarchy. In sentences like:

\pex
\a Will the boy who is crying \_ go outside? \hfill (correct inversion)
\a *Had the boy who is crying will go outside? \hfill (incorrect embedded inversion)
\xe

the auxiliary must be fronted from the main clause—not an embedded one. Human speakers never front embedded auxiliaries, even if they are linearly first.

GPT-4 replicates this behavior flawlessly, with 100\% accuracy in both English and Norwegian. LLaMA performs similarly in English and slightly less so in Norwegian, with most deviations only one point off. This aligns with earlier findings (e.g., \cite{ziv2025largelanguagemodelsproxies}) that show LLMs respect subject boundaries in auxiliary movement. Our results strengthen that observation, demonstrating that models not only produce well-formed inversions but also rate unlicensed fronting as unacceptable.

Notably, these distinctions cannot be captured by n-gram statistics alone. The fact that embedded auxiliaries are linearly earlier, yet never fronted, suggests that models track constituent domains. In Norwegian, where verb movement is more flexible, GPT-4 still distinguishes acceptable inversion patterns from ungrammatical ones, further supporting the hypothesis that these models induce a notion of subjecthood and phrasal structure.

Taken together, these results suggest that LLMs form internal generalizations that are sensitive to syntactic constituency, even in constructions where linear order is misleading.

\section{Discussion}

Our results suggest that LLMs encode abstract syntactic structure as a property drawn out of predictive training. In both inversion and parasitic gap constructions, the models reliably prefer grammatical forms mirroring human intuitions,

This supports the view that grammar can be learned from form alone. Attention mechanisms likely induce notions of subjecthood and constituency, aligning with the idea of syntax as an emergent attractor, as described in \cite {ravishankar2021attention}. By treating LLMs as proxies \cite{wilcox2022predictive}, we gain a new empirical lens on what linguistic patterns are learnable from input distributions.

Grammaticality judgments together with their corresponding utterances, can be recast as epistemic objects shared between human theory and model behavior \cite{rheinberger1997epistemic}. That models capture such distinctions without symbolic rules strengthens the case for structure as learnable rather than innate.

\paragraph{Masked-LM vs.\ autoregressive training.}
Our results may also help clarify why autoregressive models outperform masked-LM architectures on structure-sensitive tasks. Masked-LM training introduces inference conditions that do not occur in natural language use, fragmenting the representational space. By contrast, autoregressive training yields a single, globally coherent state over the entire input. This may facilitate the emergence of hierarchical generalizations—such as subjecthood, clause boundaries, and dependency licensing—independently of the specific training objective.

\section{Conclusion}

LLMs demonstrate sensitivity to structural constraints (e.g., verb inversion, parasitic gap licensing) recoverable from training data, supporting constituent structure as an emergent regularity rather than an innate system. 

At the same time, models perform less consistently on constructions like ATB extraction, suggesting that not all aspects of grammar are equally accessible. Some patterns probably depend on discourse or semantic features, beyond what surface data provides.

Rather than asking whether LLMs “know” grammar, we ask: which structural patterns are learnable from data? The answer, we argue, supports a view of grammar as an emergent regularity shaped by exposure and general formation principles in contrast to a predefined innate system.

\section*{Data and Code Availability}
All data, prompts, and scripts are available at:  
\url{https://github.com/Yoonsen/CHR-2025}\\

\bibliography{bibliography}

\end{document}